# Sliding window approach based Text Binarisation from Complex Textual images

Chitrakala Gopalan (Corresponding Author)
Dept. of Computer Science and Engineering
Easwari Engineering College, Anna University
Chennai, India

D.Manjula
Dept. of Computer Science and Engineering
College of Engineering, Anna University
Chennai, India

*Abstract*— Text binarisation process classifies individual pixels as text or background in the textual images. Binarization is necessary to bridge the gap between localization and recognition by OCR. This paper presents Sliding window method to binarise text from textual images with textured background. Suitable preprocessing techniques are applied first to increase the contrast of the image and blur the background noises due to textured background. Then Edges are detected by iterative thresholding. Subsequently formed edge boxes are analyzed to remove unwanted edges due to complex background and binarised by sliding window approach based character size uniformity check algorithm. The proposed method has been applied on localized region from heterogeneous textual images and compared with Otsu, Niblack methods and shown encouraging performance of the proposed method.

*Keywords*- Binarization; Character segmentation; textured background; Iterative thresholding; Edge spread analysis.

## I. INTRODUCTION

Understanding texts from natural scenes/scene text image such as, commercial signboards, traffic signs, and advertising billboards is very useful in many purposes such as assistant system for impaired persons, drawing attention of a driver to traffic signs, text translation system for foreigners, potential applications like license plate recognition, digital note taking, document archiving and wearable computing. Binarization problem concerns classifying individual pixels as text or background. Binarization is necessary to bridge the gap between localization and recognition by OCR. The output of this step is a binary image where black text characters appear on a white background. Current techniques are categorized into two groups: global binarization and local binarization or adaptive binarization.

In global binarization methods [12], global thresholds are used for all pixels in image and are not suitable for complex and degraded document images. Global methods are fast and robust for small text. In the other hand, local binarization methods change the threshold adaptively over the image according to properties of local regions. Local binarization methods are proposed to overcome binarization drawbacks in global ones. Local binarization methods can be improved by calculating local thresholds within separate windows or areas [11] [5]. In most of these methods, the size and shape of the window are predefined parameters. Poor binarization results are obtained when a window's boundaries cross characters and may give rise to broken characters and voids, which may cause undesirable artifacts in the binary image.

And other challenging issue related to binarising text from textual images is the presence of complex/textured background. Here sliding window approach based binarisation method is proposed to binarise the text from color documents with textured /complex background. The paper is organized as follows. Section 2 deals with prior and related works, Section 3 illustrate our method with various modules in Section 4-6. In Section 7, experimental results are reported and conclusions and future works are summarized in Section 8.

## II. RELATED WORKS

Various text binarization techniques have been found in literature and are discussed here. Dual binarization method is proposed in [1] which can easily segment texts with different two color polarities from backgrounds in the key caption area. [10] Proposed a binarisation method to remove the background pixels inside the characters also. The final binary image is gotten by fusing the three binary images such as the locally adaptive seed-fill method, the locally adaptive thresholding method and the stroke-model-based method. A learning-based binarization method is proposed in [3] for same-type documents. In this paper, the stroke width is used to evaluate the binarization .This approach can be used for only same type of documents. In [4] technique based on a Markov Random Field (MRF) model of the document is proposed. The model parameters (clique potentials) are learned from training data and the binary image is estimated in a Bayesian framework.

In [5] Segmentation of text in the detected text region is performed with all color components into two distinctive colors to discriminate between text and other non-text region with fuzzy c-mean (FCM) clustering to depict the color distribution. Adaptive local thresholding based on a verification-based multi threshold probing scheme is proposed in [6]. This approach is regarded as knowledge-guided adaptive thresholding. [7] proposed a text segmentation method based on spectral clustering and the histogram of intensity is used for the object of grouping. This algorithm uses the normalized graph cut measure as the thresholding principle to distinguish an object from the background.

However these methods mainly work on the images of text with nearly uniform background. Some works deal with complex background [8] [9] [13] and binarization applied on single character [9] and on each word [8]. But they assume the text color to be uniform and not suitable for color documents with multicolored text. [2] deals with multicolored text documents regardless of the polarity of foreground-background shades with edge-box analysis. However, if the background is textured, the edge components may not be detected correctly due to edges from the background and edge-box filtering strategy fails.

Therefore, it is proposed to address the above issues by proposing the Sliding window based character size uniformity check algorithm to minimize the complexity of background.

### III. SYSTEM DESCRIPTION

Here, an approach is proposed in which iterative thresholding is used to detect edges instead of fixed global threshold which will usually work well for images with uniform background, but not for textured background. Here uniformity of character sizes is analyzed to remove false edges due to textured background. Proposed binarization technique consists of the following processes: Preprocessing, Iterative thresholding for edge detection, Edge box formation, False Edge box removal and binarization by Sliding window algorithm as shown in Fig.1.

Suitable preprocessing techniques are applied here first to increase the contrast of the image and blur the background noises due to textured background. Then edges are detected with iterative thresholding and bounding box is generated for the detected edges. Then false edge boxes are removed and image is binarised by checking the uniformity of character box sizes.

### IV. PREPROCESSING

The main objective of the preprocessing step is to make the foreground objects more clear than the background so as to help further edge detection stage to give candidate text edges clearly. Here, original color image is transformed to a grey level image and followed by contrast enhancement based on entropy calculation, conditioning by smoothing and grey scale extension [17].

#### A. Conditional Contrast Enhancement

Images taken under a poor lighting condition may result in low entropy, which needs increase in contrast of the image for better processing. Therefore, entropy can be used as an indication if an increase in the contrast of the image will be necessary. If the entropy calculated of the image is too low, the contrast can be increased, otherwise the detected edges of characters may not form closed shapes. Entropy can be computed as follows:

$$H = -\sum \leq p_i \times \log\ p_i = \sum \leq p_i \times \log\ \frac{1}{p_i} \quad (1)$$

$H$ denotes the entropy of an image; $p_i$ represents the proportion of greyscale values in the range $i \in [0.255]$ over the entire image. From extensive experiments, an empirical value of $H_{thres} = 38$ is recommended to identify images, which needs to be enhanced in terms of the contrast.

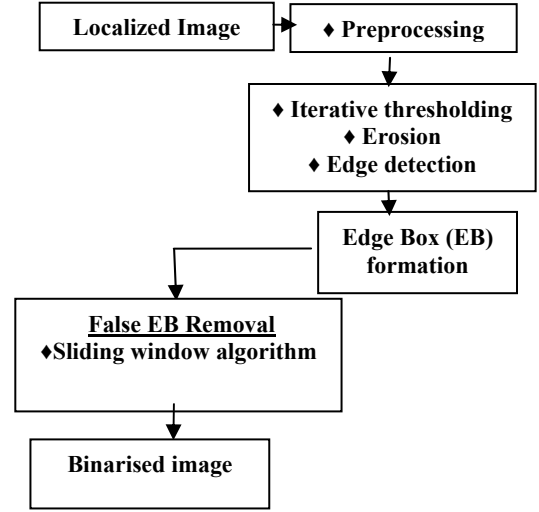

Fig 1. System Architecture of the proposed method

Based on this threshold, the contrast of the image needs to be increased as follows:

$$C(x,y) = \frac{255}{1 + \exp\left(\frac{avg\_T - T(x,y)}{v}\right)} \quad (2)$$

$C(x, y)$ denotes the transformed results. $avg\_T$ is the average grey values in the image represented by $T$. The parameter $v$ can be fixed at $v = 15$.

#### B. Smoothing

Then a simple smoothing/blurring process is applied to minimize the effects of these disturbances on the edge detection process and grey scale extension is performed for the conditioned image to increase further contrast.

$$S(x,y) = \frac{1}{10}(C(x,y) \oplus M) \quad (3)$$

Where $C(x, y)$ and $S(x, y)$ denote the greyscale value at position $(x, y)$ of the image before and after the transformation. $M$ represents the mask [17].

#### C. Grey scale extension

Maximum and minimum greyscale values of the smoothed image can be computed. If $max\_S$ and $min\_S$ are close to each other, e.g., $max\_S - min\_S < 80$, the greyscale image will be monotonous with a low contrast. Greyscale extension will increase contrast via exploiting the full range $max\_S - min\_S = 255$. Transformation equation is given as,

$$G(x, y) = (S(x, y) + \alpha) \times \beta \quad (4)$$

Where $S$ is the grayscale image after the smoothing transformation and $G$ represents the image following extension.

$$\alpha = -\min\_S, \quad \beta = \frac{255}{\max\_S - \min\_s}$$

Processed image after all these steps with input image are shown Fig. 2a to 2d.

## V. ITERATIVE THRESHOLDING AND EDGE DETECTION

Now edges are to be detected from the preprocessed image. Conventional edge detection algorithms belong to the high pass filtering, which are not fit for complex background images. Fixed global threshold will usually work well for images with uniform background, but not for textured background and the contrast of objects varies within the image .In such cases, it is convenient to use a threshold gray level that is a slowly varying function of position in the image[14]. And also edge continuity to be maintained and edge overlapping to be avoided. In this aspect, an iterative thresholding algorithm and morphologic erode algorithm [15] are used here to detect the edges. Following steps are used to detect edges in iterative thresholding,

1. Initial threshold $T^0, T^0 = \{T^k | k = 0\}$ is calculated using minimal and maximal gray value of the image.

$$T^0 = \frac{Z_{\min} + Z_{\max}}{2} \quad (5)$$

2. The image is segmented into two parts (R1 and R2) by the threshold $T^k$

$$R_1 = \{f(x, y) | f(x, y) \geq T^k\} \quad (6)$$
$$R_2 = \{f(x, y) | 0 < f(x, y) < T^k\} \quad (7)$$

3. Average gray level value of R1 and R2 is calculated and denoted as Z1 and Z2 separately.

$$Z_1 = \frac{\sum_{f(i,j) < T^k} f(i,j) \times N(i,j)}{\sum_{f(i,j) < T^k} N(i,j)} \quad (8)$$

$$Z_2 = \frac{\sum_{f(i,j) > T^k} f(i,j) \times N(i,j)}{\sum_{f(i,j) > T^k} N(i,j)} \quad (9)$$

Where f(i ,j) is the gray level value of point (i,j) in the image, N(i,j) is the power coefficient of point (i,j) and it equals to 1.0 commonly.

4. New threshold $T^{k+1}$ is calculated as average of Z1 and Z2.

$$T^{k+1} = \frac{Z_1 + Z_2}{2} \quad (10)$$

5. The algorithm is completed if $T^k = T^{k+1}$ otherwise let k=k+1 and go back to step 2.

Then morphological erosion is applied which ensures that the detected edges are only one single pixel wide and makes the detection more accurately. Meanwhile, this avoids the edge overlapping caused by the increase of the edge width. After the erosion for the image the edges are extracted in the image. Because the erosion operator eliminates all the boundary points from an object and the boundary points are all one single pixel wide. If the eroded image is subtracted by the original image, the result will be one single pixel wide edge. Fig.2e, f and g show the result of the edge detection after iterative thresholding and erosion. An 8-connected component labeling follows the edge detection step and the associated bounding box information is computed. Each component, thus obtained, is termed as an edge-box (EB) as shown in Fig. 3a.

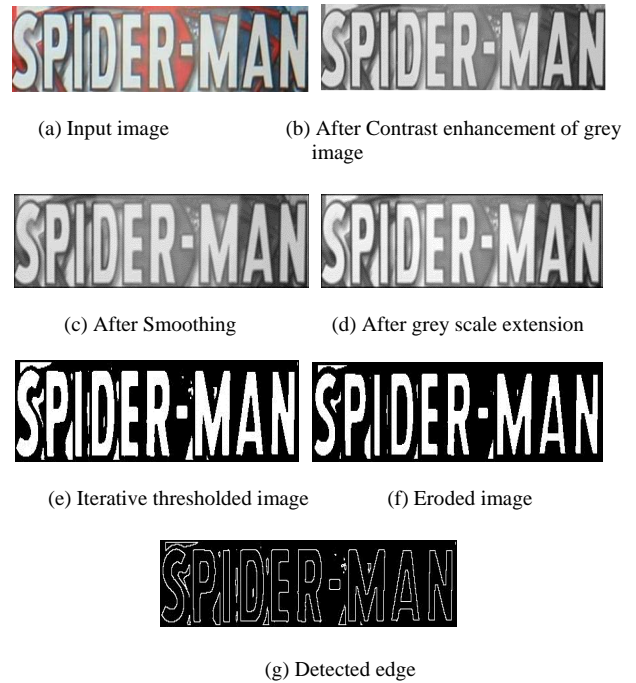

(a) Input image  (b) After Contrast enhancement of grey image

(c) After Smoothing  (d) After grey scale extension

(e) Iterative thresholded image  (f) Eroded image

(g) Detected edge

Fig 2. Edge detection

## VI. FALSE EDGE REMOVAL AND BINARISATION BY SLIDING WINDOW APPROACH

Now Each EB is checked for various characteristics of individual text character (Tamil and English) and EBs not fulfilling the criteria are removed. The following characteristics are considered for individual text character to filter out non text EBs:

1. The aspect ratio is constrained to lie between 0.1 and 10 to eliminate highly elongated regions.

2. If a particular EB has exactly one or two EBs that lie completely inside it, the internal EBs can be conveniently ignored as it corresponds to the inner boundaries of the text characters. On the other hand, if it completely encloses three or more EBs, only the internal EBs is retained while the outer EB is removed as such a component does not represent a text character. Thus, the unwanted components are filtered out by subjecting each edge component to the following constraints [2] :

   if ($N_{int}$ <3)
       {Reject $EB_{int}$, Accept $EB_{out}$}
   else
       {Reject $EB_{out}$, Accept $EB_{int}$}     (11)

### A. Sliding Window approach based false EB removal and Binarisation

Characters will have approximately uniform size in the localized region. This property is used here to remove non character EBs with the proposed Sliding window approach based binarisation. Steps involved in the algorithm are as follows:

a) Size of various EBs are calculated and store in array A [ ].

b) Initially window over the array elements is assumed by covering first two elements A[i] as a left pointer (LP) and A [i+1]) as a right pointer (RP).

c) If the difference between LP and RP is greater than threshold, consider only RP and make it as LP and slide the window size to cover next element and check the difference between those two elements.

d) Compare left end marker with remaining elements one by one and if the difference < th, expand window by including those elements into the window and freeze the window once the condition fails. This window contains identified uniform sized character EBs between LP and RP.

e) Then compute minimum of the elements within that first freezed window ($M_{W1}$ ). Here, minimum value is chosen so as to allow mixed sized characters.

f) Then slide the window to check the remaining elements and determine other valid windows ($W_i$) and minimum value of those windows as $M_{W1}, M_{W2}, ... M_{Wi}$

g) Compute minimum of $M_{Wi}$ which gives uniform edge box size as the threshold $T_s$.

3. Now binarise the image with threshold ($T_s$). EBs with size lesser than $T_s$ are removed. Foreground text pixels are shown as white and background as black pixels.

EB = 1 {If size (EB) >= Ts}
EB = 0 {If size (EB) <= Ts}     (12)

Processed image after these steps is shown in Fig 3a.

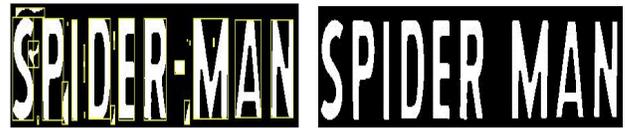

Fig 3(a) Edge box    (b) Binarised image

### VII. RESULTS AND DISCUSSION

In this section, the results of the proposed algorithm are presented. The performance of the proposed method is compared with two well-known thresholding methods, including Otsu [12] and Niblack [11] algorithms. We show here some examples without OCR comparison. We believe a visual evaluation is sufficient for a qualitative estimation of our method.

The results show that the proposed method is an effective method and outperforms the other methods in the complex / textured background situation. Disturbances due to the textured background are well avoided in our sliding window method than the compared ones. So that binarised image can be better recognized by OCR.

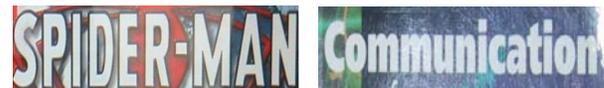
(a)    Input image

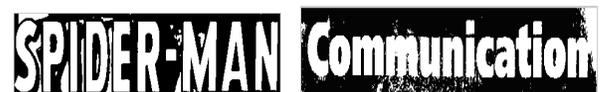
(b)    Otsu method

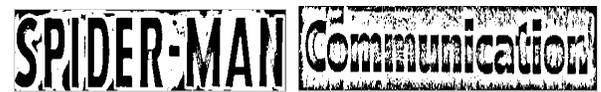
(c)    Niblack method

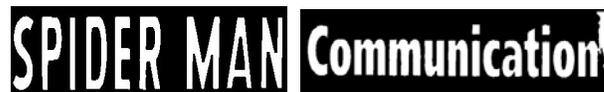
(d)    Proposed Sliding window method

Fig 6 Performance Comparison of the proposed method with other methods

## VIII. CONCLUSION

Here a novel method is proposed to binarize text from color images with textured background by analyzing character and non character edges. Sliding window based method is proposed to identify the character edges by suppressing the unwanted non character edges. Experimental results are showing encouraging performance of the proposed method with the compared binarization algorithms.

AUTHORS PROFILE

**Chitrakala Gopalan** received the B.E. and M.E degree from Univ. of Madras,India in 1995 and 2002 respectively. From 1995 to 1998 she worked as a lecturer in Raja rajeswari Engineering college ,Tamil nadu , India . From 1998 to 2002 she worked as a lecturer in Easwari Engineering college and Since 2002 , she is working as an Assistant Professor in computer science and Engineering, at Easwari engineering college, Tamil nadu , India. Currently she is Pursuing Ph.D in Anna university, Chennai, India in the department of computer science and Engineering. Her research areas of interest are Digital Image processing, Data mining and Natural language processing.

**Dr.D.Manjula** gained her B.E degree from Thiyagarajar College of Engineering in 1983 and M.E & Ph.D degree from Anna university , Chennai in 1987 & 2004 respectively. She is an Assistant professor in Department of Computer science and Engineering, Anna university Chennai, Tamil nadu,India. At present , she teaches and leads research towards language technologies , Data mining , Text mining , Imaging and networking.